\documentclass[conference]{IEEEtran}
\IEEEoverridecommandlockouts
\usepackage{booktabs}
\usepackage{amsmath,amssymb,amsfonts}
\usepackage{algorithmic}
\usepackage{graphicx}
\usepackage{textcomp}
\usepackage{xcolor}
\usepackage{mdwlist}  
\usepackage{stfloats}
\usepackage[backend=biber, sorting=none, style=ieee, maxnames=5]{biblatex}
\usepackage{siunitx}
\usepackage{makecell}
\usepackage{hyperref}
\hypersetup{colorlinks=true, breaklinks=true, 
  urlcolor=blue, linkcolor=blue,anchorcolor=blue,citecolor=blue,
  hypertexnames=true, final=true,
  pdfpagemode = UseNone, 
  pdfauthor = {},
  pdftitle = {},
  pdfsubject = {},
  pdfkeywords = {}
}

\def\BibTeX{{\rm B\kern-.05em{\sc i\kern-.025em b}\kern-.08em
    T\kern-.1667em\lower.7ex\hbox{E}\kern-.125emX}}

\begin{document}

\title{Training slow silicon neurons to control extremely fast robots with spiking reinforcement learning}
\author{
  \IEEEauthorblockN{Irene Ambrosini$^{1,2}$, Ingo Blakowski$^{1,3}$, Dmitrii Zendrikov$^{1}$,   Cristiano Capone$^{4}$, Luna Gava$^{2}$,\\ Giacomo Indiveri$^{1}$, Chiara De Luca$^{1,5, *}$, Chiara Bartolozzi$^{2, *}$}
  \IEEEauthorblockA \footnotesize$^{1}$Institute of Neuroinformatics, UZH and ETH Zurich, Switzerland\\
  $^{2}$Istituto Italiano di Tecnologia, Genoa, Italy
  $^{3}$Technical University of Munich, Munich, Germany \\
  $^{4}$ Natl.\ Center for Radiation Protection and Computational Physics, Istituto Superiore di Sanità, 00161 Rome, Italy \\
  $^{5}$ Digital Society Initiative, University of Zurich, Zurich, Switzerland \\
  $^{*}$These authors contributed equally
  }
\maketitle

\begin{abstract}
Air hockey demands split-second decisions at high puck velocities, a challenge we address with a compact network of spiking neurons running on a mixed-signal analog/digital neuromorphic processor. By co-designing hardware and learning algorithms, we train the system to achieve successful puck interactions through reinforcement learning in a remarkably small number of trials. The network leverages fixed random  connectivity to capture the task's temporal structure and adopts a local e-prop learning rule in the readout layer to exploit event-driven activity for fast and efficient learning. The result is real-time learning with a setup comprising a computer and the neuromorphic chip in-the-loop, enabling practical training of spiking neural networks for robotic autonomous systems. This work bridges neuroscience-inspired hardware with real-world robotic control, showing that brain-inspired approaches can tackle fast-paced interaction tasks while supporting always-on learning in intelligent machines.
\end{abstract}


\section{Introduction}
\begin{figure*}[t] 
\centering
\includegraphics[width=\textwidth, height=0.28\textheight, keepaspectratio]{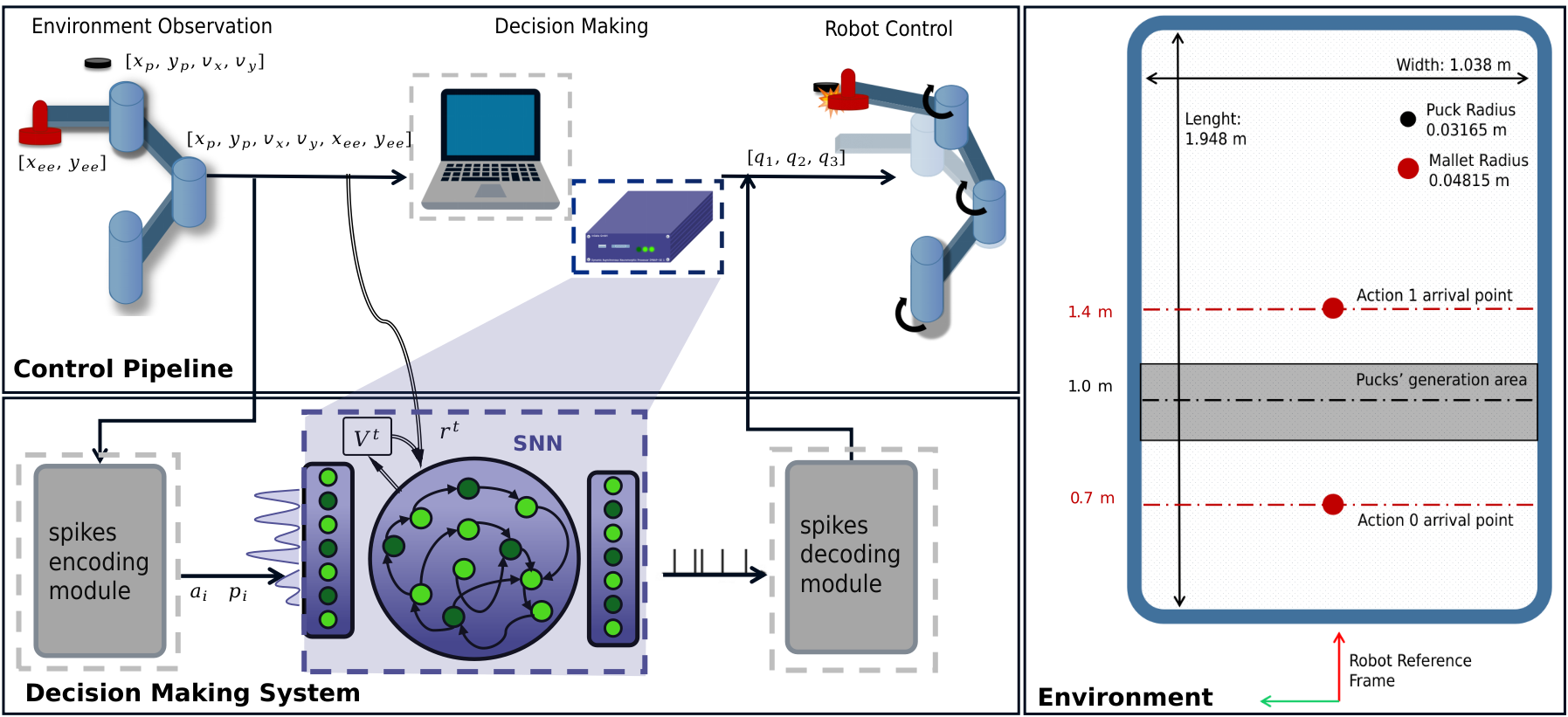}
\caption{\textbf{Control pipeline and environment.} \textbf{Top-left:} High-level flow from MuJoCo (puck $[x_p,y_p,v_x,v_y]$ and end-effector $[x_{ee},y_{ee}]$) through the decision module to the robot controller. The CPU encodes sensory data into spike trains, processed by DYNAP-SE' silicon neurons, then decoded into discrete motion primitives and translated to joint commands $[q_1,q_2,q_3]$. \textbf{Bottom-left:} Encoding module, reservoir, and spike-based readout with learning. \textbf{Right:} Table dimensions (\SI{1.038}{m} $\times$ \SI{1.948}{m}), puck/mallet radii, puck generation region, and two arrival points for Action~0/1 in the robot frame.}

\label{fig:pipeline}
\end{figure*}
\noindent The AI revolution has achieved unprecedented success by scaling neural networks to billions of parameters. Yet this progress comes at an unsustainable cost: training large models can consume megawatt-hours of energy~\cite{strubell2019energy}, while even inference on edge devices drains batteries in hours~\cite{cai2020once}. This power-hungry paradigm conflicts with the demands of autonomous robotics, where mobile platforms must learn and adapt continuously under severe energy constraints, often with budgets measured in milliwatts, not watts~\cite{floreano2015science}.Nature offers a compelling alternative. The human brain performs remarkable motor learning with roughly 20\,W of power and surprisingly few training examples~\cite{wolpert2011principles}. This efficiency gap has motivated both algorithmic innovations in sample-efficient learning and hardware advances in neuromorphic computing. On the algorithmic front, model-based reinforcement learning has demonstrated the ability to learn complex behaviors from limited data by building predictive world models~\cite{hafner2020mastering}. For instance, MuZero achieved superhuman performance in Atari, Go, Chess, and Shogi by planning with learned models~\cite{schrittwieser2020mastering}, while EfficientZero mastered Atari using only two hours of gameplay—$100\times$ less data than standard approaches~\cite{ye2021mastering}. World models have also shown robustness to model inaccuracies, while universal successor features~\cite{borsa2018universal} enable generalization to novel tasks without retraining. On the hardware front, brain-inspired neuromorphic processors process information through asynchronous spikes rather than clock cycles, achieving $1000\times$ energy reductions for specific tasks~\cite{schuman2022opportunities}. Recent chips such as Intel's Loihi~\cite{davies2018loihi}, IBM's TrueNorth~\cite{akopyan2015truenorth}, and DYNAP-SE~\cite{moradi2018scalable} have reached maturity that enables deployment in proof-of-concept robotic setups. A key step has been the introduction of always-on learning in closed-loop setups, enabling \emph{reinforcement learning} directly in the spiking domain~\cite{Khacef_etal23}. Capone et al.~\cite{capone2024towards} demonstrated biologically plausible “awake” and “dreaming” RL phases on Atari Pong, achieving competitive performance. This was later extended to real-time hardware on DYNAP-SE~\cite{Blakowski_etal2025}. At the core of these advances are \emph{spiking neural networks} (SNNs), which model neurons as leaky integrate-and-fire units that communicate through discrete spikes—a key mechanism for energy efficiency and temporal coding~\cite{johansson2004first}. Deep RL algorithms like DQN and TD3 have been implemented in SNNs~\cite{tang2021deep, akl2023toward}, but existing training methods often rely on non-local learning rules, making them computationally expensive and biologically implausible. Recent advances in local plasticity~\cite{ kheradpisheh2020temporal, goltz2021fast} enable online learning in recurrent SNNs suitable for neuromorphic hardware. However, neuromorphic RL demonstrations remain confined to simplified benchmarks—mainly 2D games with low-dimensional observations, discrete actions, and forgiving timing. Whether such systems can scale to higher-dimensional states, continuous-time control, and tight temporal constraints typical of real-world robotics remains an open question. This paper provides the first affirmative answer by implementing neuromorphic RL for robotic air hockey—a canonical fast-paced manipulation benchmark requiring predictive control under tight temporal constraints. Air hockey presents key challenges absent from game benchmarks: (1)~\emph{Higher dimensionality}: 6D continuous state space including puck position, velocity, and striker coordinates across a $1.038\times1.948$\,m workspace; (2)~\emph{Physical constraints}: precise end-effector control with mechanical latencies; (3)~\emph{Temporal dynamics}: puck trajectories at 1.0--1.5\,m/s demand predictive planning with 50\,Hz control and 1\,ms physics integration.Our work complements efforts in neuromorphic robotics addressing other layers of autonomy. Romero et al.~\cite{Jorg} demonstrated event-based vision and spiking CNNs on SpiNNaker for a 5-bar planar manipulator on a mini air-hockey table. While their approach targets fast sensorimotor integration for tracking, ours tackles the complementary problem of \emph{learning control policies through interaction}. Instead of relying on mirroring strategies, we employ neuromorphic RL to adapt behavior online. Our platform uses an anthropomorphic arm on a standard air-hockey table (Figure~\ref{fig:pipeline}), introducing larger workspaces and higher kinematic complexity. Together, these contributions point toward neuromorphic control architectures coupling efficient event-driven perception with adaptive decision-making.
\subsection{Our approach and contribution}
We extend the neuromorphic RL framework of Blakowski et al.~\cite{Blakowski_etal2025}, originally demonstrated on Atari Pong, to the far more challenging domain of physical robot manipulation. Our contributions address three key scalability challenges:
\begin{itemize*}
    \item \textbf{Beyond toy RL}: We move from simple grid-based games~\cite{brockman2016openai} to real-world air hockey with adaptive-precision continuous state encoding, bridging the gap between toy problems and embodied control.
    \item \textbf{From discrete to continuous control}: We demonstrate neuromorphic RL for continuous motion primitives executing ballistic trajectories at 50\,Hz, requiring predictive rather than frame-level decisions.
    \item \textbf{Generalization under uncertainty}: By randomizing puck positions and velocities (1.0--1.5\,m/s), we achieve 96--98\% success over 2000 episodes, showing robust learning and adaptation.
\end{itemize*}

\noindent Our system achieves robust gameplay, demonstrating that neuromorphic RL can scale from gaming benchmarks to physical robotic platforms with meter-scale workspaces and millisecond-precision constraints.

\section{Methods} \label{sec:methods}

\noindent Our approach integrates hardware-in-the-loop (HIL) learning with a simulated robotic task. Specifically, we utilize the DYNAP-SE analog-digital mixed-signal neuromorphic chip~\cite{Moradi_etal18} for closed-loop, low-latency inference at every control time step, enabling real-time interaction with the environment. This methodology allows for the rigorous testing of the neuromorphic hardware's capabilities in a dynamic, closed-loop robotic control application.

\subsection{Experimental Setup}

\noindent Experiments are conducted in the MuJoCo implementation of the \emph{Air Hockey} environment\footnote{Air Hockey Challenge: \url{https://air-hockey-challenge.robot-learning.net/}}. The setup consists of a planar air-hockey table and an anthropomorphic manipulator controlling a mallet-shaped end-effector. The agent observes the puck's 2D position and velocity $(x, y, v_x, v_y)$ and its end-effector position $(x_e, y_e)$, and must intercept the puck as it slides horizontally across the table. The main environment, robot, and control parameters are summarized in Figure \ref{fig:pipeline}.t The control loop operates at $\mathbf{\SI{50}{Hz}}$. At each step, the agent selects one of two discrete actions corresponding to pre-defined motion primitives. These primitives are executed as open-loop trajectories (using spline velocity profiles) in task space, effectively driving the end-effector toward fixed intercept points along the table center line via an embedded inverse kinematics solver, regardless of the current joint configuration. The target coordinates for the two actions: Action 0 (Home target) and Action 1 (Forward target), as detailed in Figure~\ref{fig:pipeline}. Successful interception relies on the precise temporal composition of these primitives. The underlying physics are integrated at a high fidelity of $\mathbf{\SI{1000}{Hz}}$, modeling frictionless puck dynamics and elastic collisions. Performance is evaluated based on the agent’s ability to hit the moving puck with the correct timing.

\subsection{Network architecture}

\noindent We adapt the architectural setup from \cite{Blakowski_etal2025} (originally for Atari Pong), expanding the input space to 6 state variables and reducing the readout to 2 actions, the architecture is represented in Figure \ref{fig:pipeline}.While the framework is similar, the continuous 6D state space of the air hockey task imposes fundamentally different requirements in dimensionality, precision, and dynamics (Table~\ref{tab:diff_abbr}). The meter-scale workspace demands millimeter-precision to accommodate the constraints of the robotic arm kinematics and predictive timing at puck speeds up to 1.5\,m/s.
\\
\noindent \textbf{- Input layer:} 60 FPGA-based spike generators. Each state variable is encoded as a population code using a Gaussian activity profile across 10 neurons. At each control cycle, spike trains are computed in real-time on the host CPU based on the current observed environment state and loaded into the FPGA generators, which then replay these deterministic spike patterns to the neuromorphic chip.
\\
\noindent \textbf{- Hidden layer:} 1020 silicon AdEx-LIF neurons with fixed random feedforward synapses from the input layer.
\\
\noindent \textbf{- Readout:} 2 linear units with activations calculated as a weighted sum of the filtered hidden layer spike trains. The hidden-to-readout matrix was plastic and updated with an e-prop-based learning rule~\cite{Bellec_etal20, capone2024towards, Blakowski_etal2025}, which approximates backpropagation through time in a biologically plausible way: weight updates depend only on local pre- and post-synaptic activity combined with a global reward signal, allowing fully online learning in recurrent spiking networks. The softmax function was applied across the readout neurons' activations and passed to the environment as probabilities of selecting each action. At each time step, the current environment state was encoded as spike rates on the chip, processed by the analog neuron dynamics for $20\,ms$ , then filtered and read out to evaluate the action.

\noindent Readout weights were updated every two trials:
\begin{equation}
\Delta w_{ik} = -\alpha \sum_t r^t \sum_{t' \leq t} \gamma^{t-t'} (\pi_{k}^{t'} - 1_{a^{t'}=k}) \bar{s}_{i}^{t'},
\end{equation}
\noindent where $\bar{s}_{i}^{t'}$ is the filtered activity of neuron $i$ at step $t'$, $\alpha$ the learning rate, $r^t$ the reward with discount $\gamma^{t-t'}$, and $\pi_{k}^{t'}$ the action probability, from which a one-hot vector for the selected action is subtracted.

\begin{table}[t]
\centering
\caption{Abridged — key differences (Atari Pong vs Air Hockey).}
\label{tab:diff_abbr}
\scriptsize
\renewcommand{\arraystretch}{1.4} 
\begin{tabular}{@{}p{0.28\columnwidth}p{0.32\columnwidth}p{0.32\columnwidth}@{}}
\toprule
\textbf{Aspect} & \textbf{Atari Pong \cite{Blakowski_etal2025}} & \textbf{Air Hockey (this work)} \\
\midrule
Environment \& dimensionality
  & 2D Atari game, 4D discrete (pixel units)
  & Planar air-hockey, 6D continuous (m, m/s) \\

Env. size \& resolution
  & $153 \times 210$ px
  & $1.038 \times 1.948$ m\\

Input encoding
  & \makecell[lt]{Pop. encoding (4x10 cells),\\ Spacing: $26.6$ px}
  & \makecell[lt]{Pop. encoding (6x10 cells),\\ Spacing: $0.1493$ m}\\

Action
  & \makecell[lt]{3 discrete (up/down/stay),\\ instantaneous}
  & 2 motion primitives, trajectories to $x_{\mathrm{home}}=0.70$ m, $x_{\mathrm{fwd}}=1.50$ m \\

Control \& timing
  & \makecell[lt]{$f_c=54.3$ / $27.2$ Hz \\ (agent-only / agent+model)}
  & $f_c=50$ Hz, $f_p=1000$ Hz \\

Network scale (I/H/R)
  & \makecell[lt]{40 / 510 / 3 (agent) \\ 43 / 510 / 5 (world model)}
  & 60 / 1020 / 2 (agent)\\

Training protocol
  & Awake + dreaming
  & Awake only \\

Empirical outcome
  & Dreaming $\Rightarrow$ $>$50\% sample-efficiency gain
  & 96–98\% success \\
\bottomrule
\end{tabular}
\end{table}

\subsection{Training reward}
\noindent The scalar reward $r_t$ is computed at each time step, and it's designed to encourage (i) producing forward puck motion and (ii) precise timing to intercept the puck. Let $x_p$ and $v_{x}$ denote puck position and $x$-velocity, and $d$ denote the Euclidean distance between the robot end effector and the puck. The reward is defined in Eq.~\eqref{eq:reward}.
\begin{equation}
\label{eq:reward}
r_t =
\begin{cases}
+20, & \text{if } x_p > 1.5\,\text{m} \land v_{x} > 0.1\,\text{m/s} \\[0em]
0.2\, v_{x}, & \text{if } d \le 0.08\,\text{m} \\[0em]
-0.1, & \text{otherwise}
\end{cases}
\end{equation}
\noindent The first case provides a large \emph{terminal reward} for successfully driving the puck past the target with forward momentum, defining an episode success. The second case provides \emph{shaping} by rewarding proximity ($d \le 0.08\,\text{m}$, obtained from the sum of puck and mallet radii, see Figure~\ref{fig:pipeline}) proportionally to the instantaneous forward puck speed ($v_x$). The small negative per-step reward imposes a time cost, encouraging the agent to minimize the episode length.

\subsection{Training and Evaluation Procedure}
\noindent One training run proceeds for $2000$ episodes per experimental run. To assess robustness to the chip's random connectivity and stochasticity, each training run was repeated with $10$ independent random seeds (i.e., independent reservoir samples). Reported learning curves display (in Fig \ref{fig:control_performance}c) the mean and the inter-quartile spread across the $10$ runs, smoothed using a moving average window of $30$ episodes. Unless otherwise stated, performance is reported as the on-line success rate (fraction of successful episodes).

\subsection{Scaling from Pong to Physical Robotics}

\noindent As detailed in Table~\ref{tab:diff_abbr}, this work scales the neuromorphic RL framework of \cite{Blakowski_etal2025} from a 2D pixel game to a physical 3D robotic task. This required increasing the network’s input dimensionality (from 4 inputs to 6), moving from discrete pixels to continuous states, and adapting control from single-step actions to composed motion primitives. These results demonstrate the viability of event-driven e-prop and reservoir architectures for low-power, predictive control in fast real-world robotics.

\section{Experimental conditions and results}
\noindent We conducted two sets of experiments to evaluate performance (Figure \ref{fig:control_performance}): task-level generalization under varying initial puck conditions (Fig. \ref{fig:control_performance}c, right) and encoding-range scalability tests (Fig. \ref{fig:control_performance}c, left).

\subsection{Task-level generalization}
\noindent We evaluated four conditions of increasing complexity, all measured in the robot frame (in Fig.~\ref{fig:pipeline}). The simplest case, a stationary puck at \SI{1.0}{m} distant from the robot reference frame, we achieved 100\% success rate within 200 trials. A constant-speed lateral launch from the table edge similarly reached 100\% success after 1000 episodes. Introducing speed variability ($v\in[1.0,1.5]$\,m/s) increased learning time, with success stabilizing above 96\% after 1500 episodes. Finally, randomizing both initial position (within a 0.10\,m window) and speed yielded the highest asymptotic performance, stabilizing above 98\% after 1300 episodes. This broader state distribution likely prevents readout overfitting and promotes generalizable closed-loop strategies.

\subsection{Encoding-range scalability}
\noindent With a fixed set of 1020 silicon neurons, we systematically varied the puck speed range to explore the representational limits. A narrow range ($[0.7,0.9]$\,m/s) achieved $\textgreater97\%$ success within $\thicksim150$ episodes, while a medium range ($[0.7,1.2]$\,m/s) required $\thicksim700$ episodes to reach similar performance. The widest range ($[0.7,1.5]$\,m/s) resulted in a modest $\sim4\%$ drop in asymptotic success (from $97\%$ to $93\%$), indicating graceful degradation. These results confirm that wider input ranges increase convergence time and slightly reduce performance, consistent with the finite resolution of a fixed-size network.

\subsection{Behavior consolidation through learning}
\begin{figure}[h]
    \centering
    \includegraphics[width= 0.60\textwidth, height=0.55\textheight, keepaspectratio]{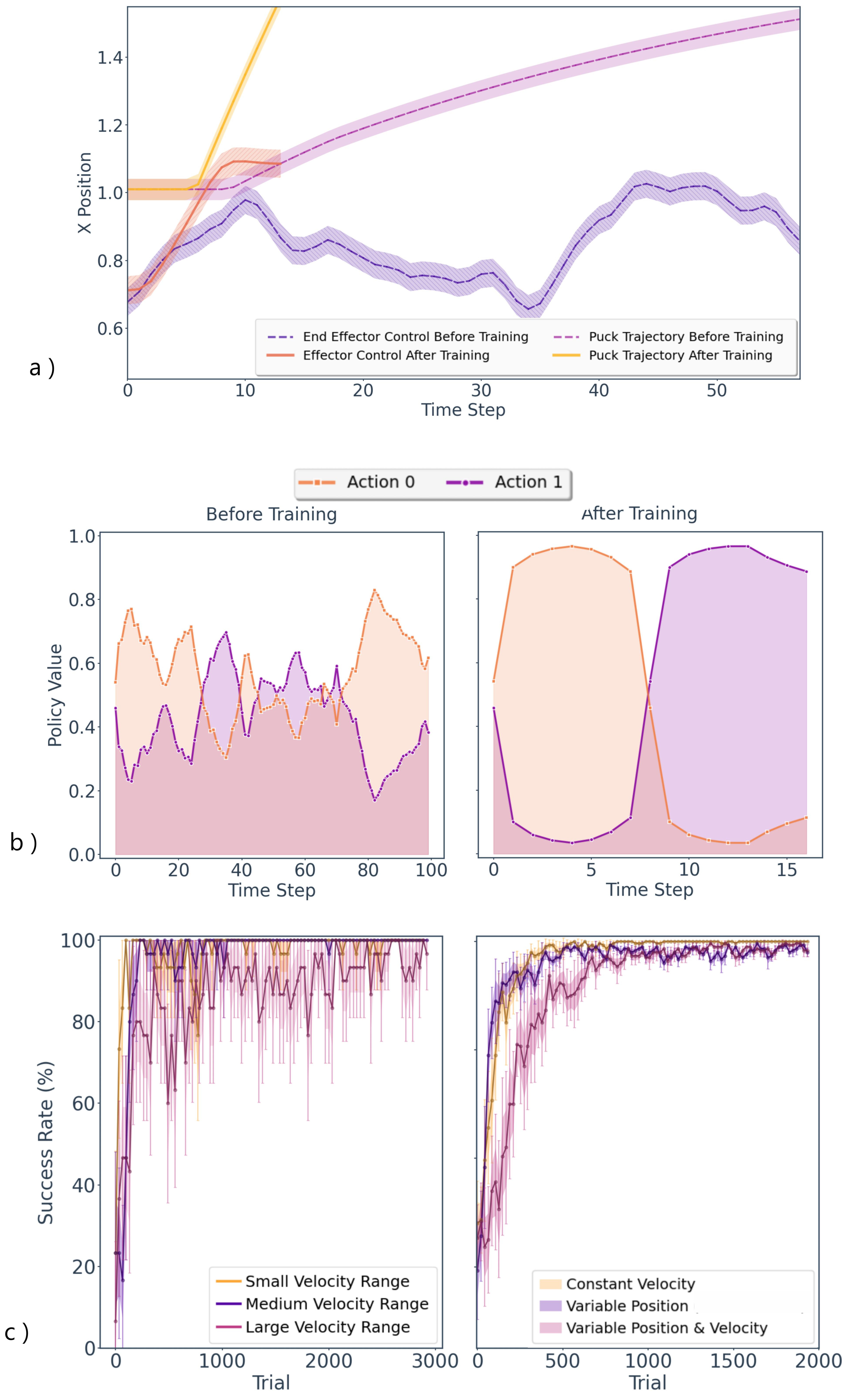}
    \caption{%
    \textbf{Neuromorphic learning masters interception timing and generalizes robustly.}
    \textbf{(a) Timing acquisition:} Pre-training (dashed) shows erratic actions; post-training (solid) achieves immediate, low-variance interceptions, reflecting learned timing.
    \textbf{(b) Policy evolution:} Stochastic switching between motion primitives consolidates into a deterministic sequence—early commitment followed by precise, tactical switching.
    \textbf{(c) Generalization:} \emph{Left:} Learning curves for three initial velocity ranges. \emph{Right:} Robustness panel comparing convergence across dynamic conditions.
    }
    \label{fig:control_performance}
\end{figure}

\noindent Before training, the agent exhibits near-random action selection with high variability (Fig.~\ref{fig:control_performance}a--b, dashed). Learning transforms this stochastic exploration into a deterministic, temporally precise strategy characterized by early commitment to the interception trajectory followed by precisely timed corrective maneuvers (solid lines). This consolidation minimizes timing variance and enables robust generalization across varied initial conditions, demonstrating the network's capacity to extract coherent, reliable policies from noisy state observations.

\section{Discussion}

\noindent We demonstrate that neuromorphic reinforcement learning can perform fast robotic manipulation using 1020 DYNAP-SE neurons, achieving stable performance within 1500–2000 episodes. Building on simulation-based bio-inspired RL for air hockey~\cite{ambrosini2024timing}, we show that neuromorphic hardware achieves better performances with 10$\times$ fewer neurons (1020 vs 10,000). This bridges the gap from gaming benchmarks to air hockey's 6D continuous state space and millisecond-precision interceptions at 1.5~m/s. However encoding-range tests reveal constraints of a fixed-size silicon network: small velocity ranges ($v\in[0.7,0.9]$~m/s) reach 97\% success, moderate ranges ($v\in[0.7,1.5]$~m/s) 93\%, and wide ranges ($v\in[0.5,2.0]$~m/s) drop to 86\%. Using all four DYNAP-SE cores would increase capacity, likely restoring performance and illustrating modular multi-core scalability. Future improvements include adopting the biologically-plausible dreaming mechanism~\cite{capone2024towards, Blakowski_etal2025} for offline learning, exploiting additional neuromorphic resources already available, or moving the readout on-chip to save energy from computing digital eligibility traces $\bar{s}_i^t$, compensating for lower resolution with a larger hidden layer. Direct event-camera input~\cite{gava2022puckparallelsurfaceconvolutionkernel} could enable native asynchronous processing with sub-ms latency, and deployment on platforms like iCub~\cite{metta2010icub} would validate robustness under real-world sensor noise and mechanical perturbations.

\noindent This work demonstrates that complex sensorimotor learning is achievable with just 1020 analog-digital silicon neurons, highlighting the potential of mixed-signal neuromorphic processors for real-time autonomous robotics.






\section*{Acknowledgments}
We acknowledge the financial support from PNRR MUR
Project PE000013 ”Future Artificial Intelligence Research
(hereafter FAIR)”, funded by the European Union – NextGenerationEU.
C.D.L has received funding from: Bridge Fellowship founded by the Digital Society Initiative at University of Zurich (grant no.G-95017-01-12).

\printbibliography
\end{document}